\documentclass[letterpaper, 10 pt, conference]{ieeeconf}  

\IEEEoverridecommandlockouts                              

\usepackage{graphicx} 
\usepackage{amsmath} 
\usepackage{amssymb}  
\usepackage{booktabs} 
\usepackage[labelfont=bf,labelsep=period,font=normalsize]{caption}
\usepackage{url}

\title{\LARGE \bf
inr2action: Encoding Robot Episodes as Neural Implicit Representations for Recall and Action
}

\author{Leann Alhashishi$^{1}$ and Christos Margadji$^{1,2}$ and Sebastian W. Pattinson$^{1}$
\thanks{$^{1}$ Institute for Manufacturing, Department of Engineering,\newline University of Cambridge    {\tt\small \{la563,cm2161,swp29\}@cam.ac.uk}}
\thanks{$^{2}$ Matta Labs, 77 East Road, London, United Kingdom }%
}

\begin{document}

\maketitle{}
\thispagestyle{empty}
\pagestyle{empty}

\begin{abstract}
Long-horizon robotic operation requires memory that scales with essential information rather than execution time. We present a parameter-space memory framework where each episode is stored as a separable object. A neural implicit representation is fitted to episode video from a shared, meta-learned initialization; the resulting weight delta constitutes the memory. Crucially, this delta remains a fixed size regardless of whether the episode spans a hundred or a hundred thousand frames. A learned projector maps the weight delta into token embeddings, conditioning a language model for question answering and a decoder for joint command regression, without frame retrieval at query time. On RoboMME, the language model exceeds chance across all question types, and the decoder achieves a mean $R^2$ of 0.51. Replacing the delta with matched noise drops performance by 11.4 points, while random-origin weights are non-decodable, proving the meta-learned initialization is essential for embedding episode content. By representing individual experiences as fixed-size weight deltas in a common parameter space, this approach decouples memory cost from time horizon and provides a basis for scalable streaming memory.
\end{abstract}

\section{Introduction}
\label{sec:intro}
\begin{figure*}[htbp]
    \centering
\includegraphics[width=\linewidth]{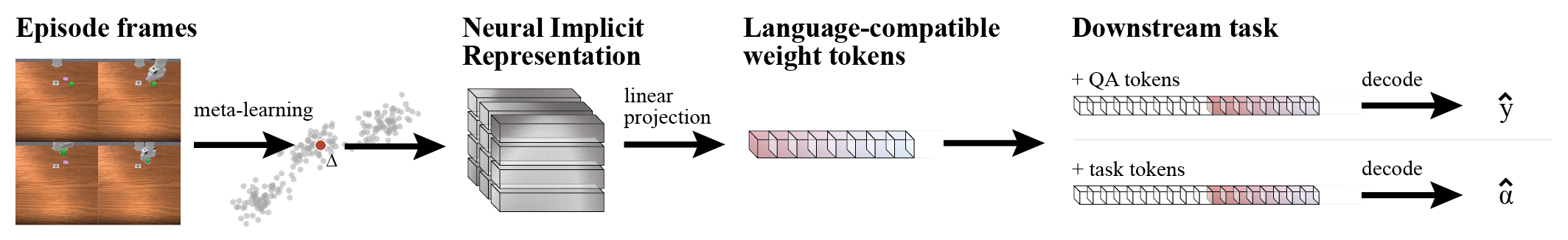}
\caption{\textbf{The overall system using weights as memory.} An episode is encoded by fitting an INR from a meta-learned initialization $\theta$; the weight delta $\Delta$ is the memory and reconstructs the frames. With no frame access, the same $\Delta$ is decoded into the next joint command and a natural-language answer.}
\end{figure*}

Robots that operate over extended horizons must remember what they have observed.
When the information that determines the correct action has already left the field of view, a robot without memory can only guess.
In robotic manipulation, the evidence that resolves such a decision can lie seconds, minutes, or in long-horizon deployments hours in the past, and by then the scene itself may no longer show it: a quantity that accumulated over time, a property visible only before it was occluded, a reference fixed at the outset of the episode.
Feeding the entire observation stream back into the policy is not a general answer, as its cost scales with how long the robot has been running rather than with how much of what it saw actually needs to be remembered.
We consider such settings non-Markovian, as the information needed at time $t$ is distributed across the episode rather than contained in any single frame, a gap that recent benchmarks make explicit~\cite{dai2026robommebenchmarkingunderstandingmemory}.
Throughout, memory refers to a persistent representation of information that can be accessed independently of the original input that produced it.

Existing sequence models hold information in one of three places.
The first is transient activations: an LSTM's hidden state~\cite{10.1162/neco.1997.9.8.1735} or a transformer's key-value cache~\cite{vaswani2023attentionneed} compresses observed history, but the stored information is bound to the current computation, and once the forward pass ends or the context window closes it cannot be queried on its own terms.
The second is explicit external storage, from differentiable memory matrices~\cite{graves2014neuralturingmachines} to retrieval over document collections~\cite{lewis2021retrievalaugmentedgenerationknowledgeintensivenlp}, which decouples storage capacity from model capacity at the cost of memory becoming a record that must be retrieved before reasoning can proceed.
A third and more recent paradigm encodes memory directly into parameters: Titans~\cite{behrouz2024titanslearningmemorizetest} updates a dedicated memory module's weights during inference, and GradMem~\cite{kuratov2026gradmemlearningwritecontext} performs test-time gradient descent on learnable memory tokens, both showing that gradient-based optimization at inference time can construct persistent memory for a model.
These methods, however, have been developed primarily for language sequences, and they continually update a single running memory over an incoming stream rather than representing individual experiences as separable memories.

In this work, each robot episode is encoded as its own parameter-space memory: a neural implicit representation (INR) fitted to the episode from a shared meta-learned initialization, whose weights constitute the memory.
A learned projector maps these weights into token embeddings for a language model, which answers natural-language questions about the episode, and the same tokens condition an action decoder that predicts the robot's next joint command.
No frame is stored, reconstructed, or retrieved at query time.
We evaluate on RoboMME~\cite{dai2026robommebenchmarkingunderstandingmemory}, a benchmark of 16 non-Markovian manipulation tasks spanning temporal, spatial, object, and procedural memory, and show that a memory encoded in this way is legible to a downstream model without frame re-access.
The contributions of this work are:

\begin{itemize}
    
    \item A formulation of episodic memory for robotics as INR weights, and a tensor-wise projector mapping them into the token space of pre-trained language backbones, yielding, to our knowledge, the first system to answer natural-language questions about a video episode from INR weights alone.
 
    \item An action decoder of the same weight-tokens that regresses the robot's optimal next joint command, extending weight-space memory from recall to action.
 
    \item Matched noise and within-task controls, together with an initialization ablation, attributing both question answering and action decoding to episode content and establishing the shared meta-learned origin as the condition that renders the weights decodable.
    
\end{itemize}

\section{Related work}
\subsection{Neural Implicit Networks}

A neural implicit representation (INR) is a neural network trained to encode a signal as a continuous function, $f(x) \to y$, mapping input coordinates directly to signal values rather than storing data on a discrete grid. Once trained, the network can then reconstruct the original signal through forward passes over the learned parameters. SIREN~\cite{sitzmann2020implicitneuralrepresentationsperiodic} established the basic feasibility of representing perceptual signals this way: its periodic sine activations counteract the spectral bias of ReLU networks, which limits the fine detail they can represent, and yield high-fidelity reconstructions of images, objects, and audio.

Subsequent work extended what fitted parameters can hold.
NeRF~\cite{mildenhall2020nerfrepresentingscenesneural} extended implicit representations from two dimensions into three, so that scene geometry and appearance could be recovered directly from learned parameters.
LeRF~\cite{kerr2023lerflanguageembeddedradiance} built on this by distilling CLIP and DINO feature embeddings into a NeRF's weight space, allowing open-vocabulary semantic queries over the encoded scene.

Video, being a substantially larger signal than a static image or scene, makes fitting speed a practical bottleneck for these pixel-wise architectures. NeRV \cite{chen2021nervneuralrepresentationsvideos} represents video as a function of time $t$, predicting a full RGB frame from a single temporal coordinate rather than querying spatial locations one at a time, which is considerably faster to fit and reconstruct than pixel-wise INRs \cite{gwilliam2025designtrainimplicitneural}. 

Fitting cost has itself been a target, falling from hours to seconds through alternative parameterizations~\cite{M_ller_2022, chen2022tensorftensorialradiancefields}, amortized prediction of weights in a single forward pass~\cite{chen2022transformersmetalearnersimplicitneural}, or meta-learned initializations that converge in a handful of gradient steps~\cite{sitzmann2020metasdfmetalearningsigneddistance}, making per-instance fitting practical online rather than offline.
This literature establishes that perceptual detail, geometry, semantics, and temporal evolution can all be stored in INR parameters, but it largely treats fitted weights as a route to reconstruction and compression rather than as representations to be queried in their own right.

\subsection{Weight Space Learning}
\label{sec:weightspacelearning}
 
Weight space learning treats the parameters of trained networks as a data modality in their own right~\cite{han2026surveyweightspacelearning}.
Functa~\cite{dupont2022datafunctadatapoint} (data as a function) introduced this framing for INRs, treating each fitted function as a data point in a learned representation space, with classifiers trained directly on these resulting modulation vectors generalizing to unseen instances.
This effect depends on all fitted functions sharing a common origin, an observation grounded via the Implicit Function Theorem~\cite{qiu2026ensuringsemanticsweightsimplicit}.
 
A shared initialization is the standard means of supplying such an origin.
MAML~\cite{finn2017modelagnosticmetalearningfastadaptation} learns an initialization from which a few gradient updates produce task-specific parameters, which offers a way to produce fitted weights that remain comparable across instances.
COIN++~\cite{dupont2022coinneuralcompressionmodalities} applies this combination of shared initialization and fast per-instance fitting to INRs specifically by meta-learning a shared base network and representing each new signal through low-dimensional modulations fitted at inference time.
It is a close precedent for treating adaptation from a common origin as an encoding mechanism.
Gielisse et al.~\cite{gielisse2025endtoendimplicitneuralrepresentations} show that meta-learned SIREN initializations combined with a transformer over the resulting weights can outperform permutation-equivariant alternatives on classification.
 
Once fitted weights sit in a shared coordinate system, standard learning algorithms can exploit that structure directly.
De Luigi et al.~\cite{deluigi2023deeplearningimplicitneural} show that an encoder trained over fitted INR weights for 3D shapes yields embeddings that support retrieval, clustering, and classification.
Weight-space representations have more recently begun serving as direct inputs to foundation models: LLaNA~\cite{amaduzzi2024llanalargelanguagenerf} projects fitted NeRF weights into the embedding space of a large language model, following the training scheme of LLaVA~\cite{liu2023visualinstructiontuning}, so that the model answers natural-language queries about a 3D scene while conditioning on weights rather than on pixels, image features, or rendered views.

\begin{figure*}[htbp]
    \centering
    \includegraphics[width=\textwidth]{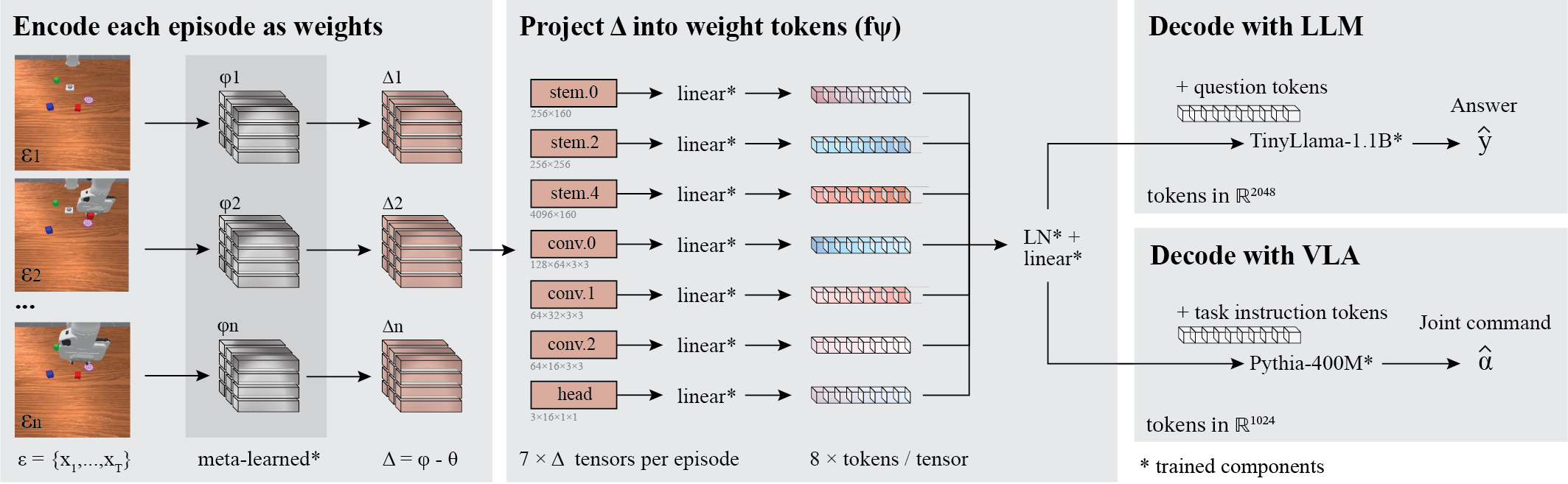}
    \caption{\textbf{Overall System Pipeline.} (1) Each episode is fitted as an INR from the shared meta-learned initialization $\theta$, yielding a weight delta $\Delta$. (2) Each tensor of $\Delta$ is projected independently and pooled into 8 tokens, giving 56 weight-tokens. (3) The tokens condition either the recall head (TinyLlama-1.1B) or the action head (Pythia-400M), with no access to frames.}
    \label{fig:system_overview}
\end{figure*}

\subsection{Memory in Robotic Manipulation}

Robot manipulation systems store interaction history in the same three places identified in Section~\ref{sec:intro}: as an external record, in activations, or, far more rarely, in parameters. The first is the most common and takes two forms. Symbolic approaches derive non-differentiable abstractions from the interaction history, tracking object-centric states~\cite{chen2026historyawarevisuomotorpolicylearning} or appending past actions to the input prompt~\cite{bu2025univlalearningacttaskcentric}, with recent systems using large vision-language models to write language-based subgoals or memory summaries~\cite{sridhar2025memerscalingmemoryrobot, geminiroboticsteam2025geminirobotics15pushing, torne2026memmultiscaleembodiedmemory}. Offloading memory and reasoning externally in this way is costly at inference and makes end-to-end optimization difficult. Perceptual approaches instead retain differentiable visual or multimodal features, either appended directly to the model input as context tokens~\cite{jang2025contextvlavisionlanguageactionmodelamortized} or cached in memory banks for later retrieval~\cite{fang2025sam2actintegratingvisualfoundation, shi2026memoryvlaperceptualcognitivememoryvisionlanguageaction}. Recurrent approaches compress history into a fixed-size latent state through iterative updates, from recurrent behavior-cloning policies~\cite{mandlekar2021matterslearningofflinehuman} to recent state-space models~\cite{mitl, liu2024robomambaefficientvisionlanguageactionmodel}. Each of these retains observations, derived features, or activation states. 

\section{Methods}

\subsection{Task definition}

Given a video episode $\epsilon$ consisting of a sequence of frames $x_1, \dots, x_T$ and a natural-language query about the episode, the system produces an output by factorizing this mapping through a weight-space memory representation. Each episode $\epsilon$ is mapped to its own parameter vector $\phi$ and, through the delta defined below, to its own memory representation $\Delta$.

An episode is first encoded into the parameters of an INR by fitting it to reconstruct the video. Let $\phi$ denote the resulting parameter vector and $\theta$ a shared and task-agnostic initialization common to all episodes. The episode's memory representation is the delta
\begin{equation}
\Delta = \phi - \theta,
\label{eq:delta}
\end{equation}
which isolates the change induced by that specific episode ($\phi$) from the initialization shared across the dataset ($\theta$). A learned projector $f_\psi$ maps $\Delta$ into a sequence of token embeddings compatible with a language model $g$. We consider two ways of querying this representation (see Figure~\ref{fig:system_overview}). For question answering, given a natural-language question $q$,
\begin{equation}
\hat{y} = g(f_\psi(\Delta), q),
\label{eq:qa}
\end{equation}
where the language model conditions on both the projected weight-tokens and the question. For action decoding, the linguistic decoding of $g$'s output is replaced by a regression head $h_\omega$ applied to its final hidden state,
\begin{equation}
\hat{a} = h_\omega\big(g(f_\psi(\Delta),\, q)\big) \approx a \in \mathbb{R}^{d},
\label{eq:action}
\end{equation}
with $q$ given by the episode's task instruction, $g$ instantiated separately per output head, and $a$ the absolute joint-space command at the point where the episode's observation ends. 

\subsection{Dataset generation}
\label{sec:dataset_generation}

All experiments use RoboMME~\cite{dai2026robommebenchmarkingunderstandingmemory}, a benchmark of question-answering videos built on the ManiSkill simulator~\cite{mu2021maniskillgeneralizablemanipulationskill} with a 7-DOF Franka Panda arm. Each video shows the arm performing a tabletop manipulation task, and the accompanying questions probe aspects of the episode that no isolated frame can resolve. The 16 task types form four suites, each targeting a distinct form of memory: Counting (temporal), Permanence (spatial), Reference (object), and Imitation (procedural).

\begin{figure}[t]
    \centering
    \includegraphics[width=\columnwidth]{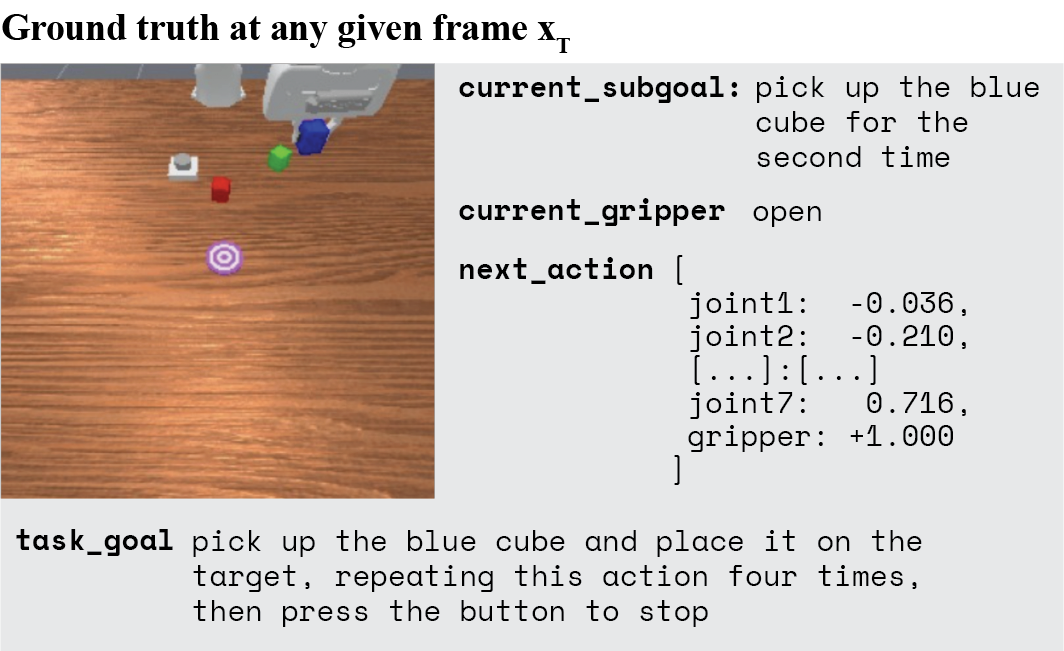}
    \caption{\textbf{RoboMME ground truth at a single frame, illustrated on an example PickXtimes episode.} Each frame is paired with the label fields read out when it serves as the cut frame $T_{\text{cut}}$: the gripper state (\texttt{gripper}), the atomic action in progress (choice\_action), the current sub-step (\texttt{subgoal}), the instruction governing the episode (\texttt{task\_goal}), and the commanded joint vector (\texttt{joint\_action}).}
    \label{fig:label_example}
\end{figure}

Each task type contributes 100 source episodes, giving a base corpus of 1,600 episodes split 70/15/15 into train, validation, and test partitions, with frames rendered at $64 \times 64$\,px. A single episode does not produce a single INR: for each episode, 10 random truncation points $T_{\text{cut}}$ are drawn from its timeline, and an INR is fit independently to the frames up to $T_{\text{cut}}$. The label fields are extracted at that last frame, so each source episode supplies multiple temporally distinct memories, for 16,000 fitted INRs per configuration (Figure~\ref{fig:label_example}). Four label types are read out at each cut, ordered by increasing dependence on memory: \texttt{gripper}, whether the gripper is closed at the cut frame, resolvable from that frame alone. \texttt{choice\_action} is the atomic action being performed. \texttt{subgoal} is a fine-grained sub-step the robot is currently executing, such as picking up a given cube for the third time, and \texttt{task\_goal}, the instruction governing the episode, whose instances within a task are near-duplicates differing in a single attribute such as a color or a repetition count.

Every INR is fit from a common initialization $\theta$, learned via MAML~\cite{finn2017modelagnosticmetalearningfastadaptation}, which optimizes $\theta$ so that a small number of gradient steps produces a good per-episode fit on any episode from the training distribution. Per-episode fitting then adapts $\theta$ through 500 steps of full-batch Adam on the reconstruction objective of Section~\ref{sec:system_architecture}. Because adapted and initialization weights are typically close in magnitude, with the initialization dominating the raw parameter values, we follow~\cite{gielisse2025endtoendimplicitneuralrepresentations} in taking the difference $\Delta$ (Eq.~\ref{eq:delta}) as the representation rather than the raw fitted weights, isolating the change induced by the specific episode. Fitting from a shared origin is also what makes deltas comparable across instances, per the shared-coordinate-system arguments of Section~\ref{sec:weightspacelearning}.

\begin{figure}[t]
    \centering
    \includegraphics[width=\columnwidth]{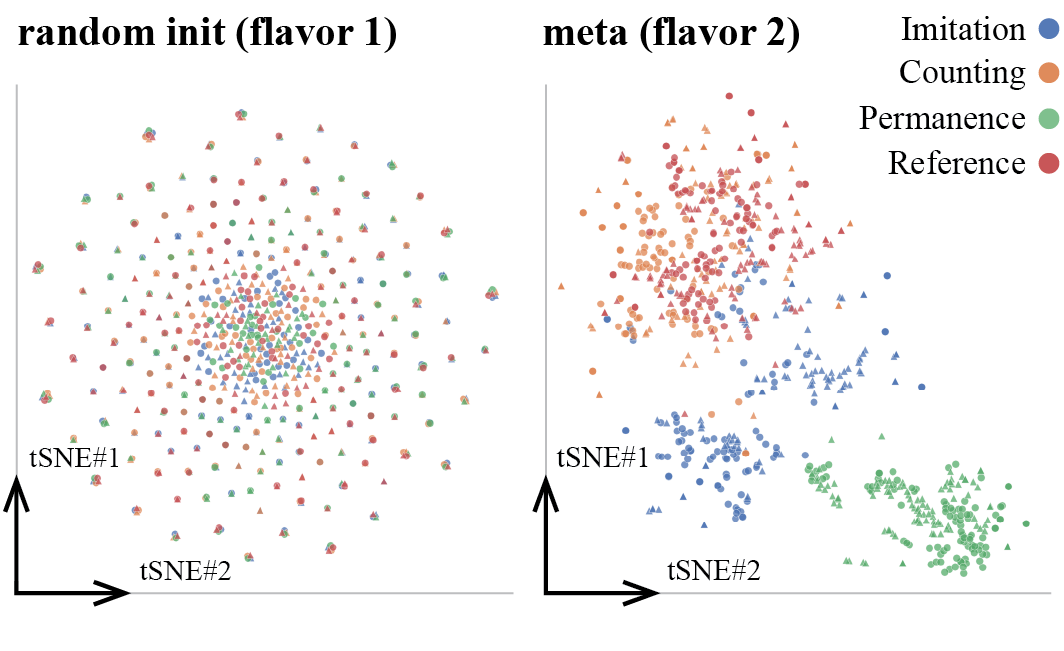}
    \caption{\textbf{t-SNE of INR weight space, meta (Flavor 2) versus random initialization (Flavor 1), colored by memory suite.} Both embeddings (PCA-100, perplexity 30) use the same 8 tasks, 2 per suite, 100 samples each. Axes are arbitrary.} 
    \label{fig:tsne}
    \vspace{10mm}
\end{figure}

To verify that this construction yields organized weight representations, t-SNE embeddings were computed over the fitted deltas (Figure~\ref{fig:tsne}). No suite or task label was used in fitting the INRs or computing the embedding. Under the meta-learned initialization, Permanence and Imitation task-type weights form tight, well-separated clusters, while Counting and Reference weights interleave with one another but remain distinct from the other two suites. Coherent clusters form even for tasks that never appeared in meta-training, which uses only five of the 16 task types. Randomly initialized weights show no visible separation by suite. This is consistent with adaptation from the shared initialization inducing an organized common weight space, and it is the delta that the downstream experiments operate on.

\subsection{System architecture}
\label{sec:system_architecture}

The system comprises three components, following Figure~\ref{fig:system_overview}: the INR encoder producing each episode's delta, the projector mapping the delta into weight-tokens, and two output heads (decoders) over the resulting token sequence.

\subsubsection{INR Encoder}
Each episode's INR follows NeRV~\cite{chen2021nervneuralrepresentationsvideos}: a positional encoding lifts the normalized timestep into a high-frequency sinusoidal embedding, a three-layer MLP stem maps it to a spatial feature map, a series of NeRV blocks progressively upsample it through a $3 \times 3$ convolution, a PixelShuffle layer~\cite{shi2016realtimesingleimagevideo}, and a GELU activation, and a final $1 \times 1$ convolution produces the RGB frame (full architectural specifications follow Appendix~A.1 of~\cite{chen2021nervneuralrepresentationsvideos}). Each INR is fit by minimizing a weighted combination of $\mathcal{L}_1$ reconstruction and structural dissimilarity,
\begin{equation}
\mathcal{L}_{\text{NeRV}} = \lambda \| \hat{x} - x \|_1 + (1 - \lambda) (1 - \text{SSIM}(\hat{x}, x)),
\label{eq:nerv_loss}
\end{equation}
with $\lambda = 0.7$: the $\mathcal{L}_1$ term encourages pixel-accurate reconstruction, while the SSIM term preserves local structural patterns that $\mathcal{L}_1$ alone can blur. During meta-training, the inner-loop learning rate is a per-parameter vector learned jointly with $\theta$ over $K = 3$ inner steps, following the MetaSDF formulation of~\cite{sitzmann2020metasdfmetalearningsigneddistance}.

\subsubsection{Weight-Token Projector}
Since $\Delta$ is a collection of tensors, one per INR weight matrix, each tensor is projected independently following the Inr2Array approach of~\cite{zhou2023neuralfunctionaltransformers}, preserving the layer-wise structure that flattening into a single vector would discard. Each tensor is treated as a set of row vectors, one row per output unit, with a convolutional tensor's input and spatial dimensions first folded into a single row. A dedicated linear map per tensor then projects every row into a shared latent width $\tilde{c} = 256$, so tensors of different shapes and roles are not forced through the same projection, yet land in the same latent space. Each tensor's tokens are pooled to a fixed $t_{\text{pl}} = 8$ per layer via adaptive average pooling, the pooled groups from the $n = 7$ active tensors are concatenated into a sequence of $N = 56$ tokens, and a final shared LayerNorm and linear map lift every token into the backbone's hidden dimension. Together, the per-tensor projections and this shared layer constitute $f_\psi$.

\subsubsection{Output Heads}
For question answering, $g$ is instantiated by TinyLlama-1.1B-Chat~\cite{zhang2024tinyllamaopensourcesmalllanguage}, whose small size keeps two-stage fine-tuning tractable. Each item's question and candidate choices are serialized into a two-turn conversation in the model's native chat template. A placeholder in the question expands into a fixed span of $N$ patch tokens whose embeddings are replaced by the projector's outputs before the forward pass, following the input construction of LLaNA~\cite{amaduzzi2024llanalargelanguagenerf}, with cross-entropy computed only on the answer tokens. For action decoding, $g$ is instantiated by the pretrained Pythia-400M~\cite{biderman2023pythiasuiteanalyzinglarge} released with TinyVLA~\cite{wen2025tinyvlafastdataefficientvisionlanguageaction}, with the vision tower, multimodal projector, and diffusion head discarded. Weight tokens are injected identically, with the question in the QA experiment being replaced by the episode's instruction. $h_\omega$, a three-layer MLP on the hidden state of the final non-padding position, regresses the eight-dimensional joint command against per-dimension normalized targets under the Huber loss~\cite{Huber1992}.

\subsubsection{Training and Implementation Details}
Both heads train in two stages, following LLaVA~\cite{liu2023visualinstructiontuning} and LLaNA~\cite{amaduzzi2024llanalargelanguagenerf}: Stage one freezes the backbone and updates only the projector and new placeholder embeddings, and Stage two fine-tunes the projector and backbone jointly at a substantially lower learning rate to limit catastrophic forgetting. Training on answer generation alone does not guarantee the model is actually using its weight-tokens, as a projector could learn to produce tokens the language model largely ignores. Each training example is therefore evaluated twice, once with its own weight-tokens, producing $\mathcal{L}_{\text{real}}$, and once with another example's weight-tokens substituted within the batch, producing $\mathcal{L}_{\text{swap}}$, and a hinge penalty
\begin{equation}
\mathcal{L}_{\text{contrastive}} = \max(0,\ m + \mathcal{L}_{\text{real}} - \mathcal{L}_{\text{swap}}),
\label{eq:swap_loss}
\end{equation}
with margin $m = 0.5$ and weight $\lambda_c = 1.0$. This is added to the objective.  Stage one trains for 15 epochs at learning rate $10^{-4}$; Stage two for 3 epochs at $2 \times 10^{-5}$ with cosine decay and 3\% warmup, extended to 10 epochs for the action readout, whose regression head trains from scratch.

\section{Experiments}
\label{sec:experiments}

All evaluations use the corpus of Section~\ref{sec:dataset_generation}. For \texttt{task\_goal} and \texttt{subgoal} questions, multiple-choice distractors are sampled exclusively from the ground-truth task's label vocabulary: pooling across tasks would let the model answer by recognizing coarse task identity alone, while restricting to the same task forces reliance on what the weight-tokens encode about the specific goal or sub-action. At evaluation, every candidate choice is scored twice, with and without the weight-tokens spliced in, and the reported score is the difference
\begin{equation}
\text{score}(c) = \text{NLL}(c \mid \text{prompt } + \text{ $\Delta$ tokens}) - \text{NLL}(c \mid \text{prompt}),
\label{eq:debias}
\end{equation}
so that predictions reflect how much the weight-tokens shift the model's preference for an answer rather than the model's own priors. The prediction is the minimum-scoring candidate. Because candidate counts vary per item, the chance baseline is the average per-item random-guess rate. Accuracy is reported as mean $\pm$ standard deviation across 2 training seeds, alongside Cohen's $\kappa$~\cite{doi:10.1177/001316446002000104}. As a further control, every configuration is re-evaluated with $\Delta$ replaced by Gaussian noise of matching shape passed through the trained projector, a substitution that removes only the information deriving from the episode itself.

\subsection{Weight-Space Question Answering}
\label{sec:qa_experiments}
\begin{figure}[t]
    \centering
    \includegraphics[width=\linewidth]{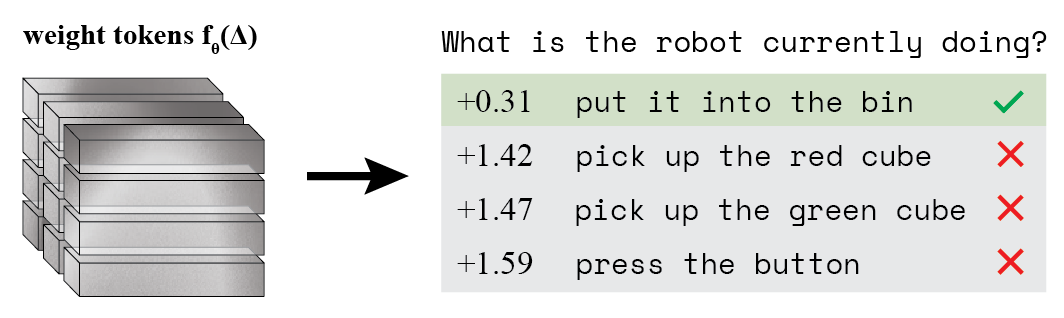}
    \caption{\textbf{Choice-scoring at inference}, illustrated on an example item from Flavor 2. Every candidate answer is scored by its prior-debiased negative log-likelihood (Eq.~\ref{eq:debias}). Scores are shown beside each candidate and the prediction is the minimum. A \texttt{subgoal} item answered correctly, with the true candidate separated from the distractors by a clear margin.}
    \label{fig:llm_example}
\end{figure}

The central result is that a language model, given 56 projected weight-tokens ($tpl$ $\times$ 7 tensor groups) and no access to any frame, exceeds its chance baseline on every question type after Stage 2 (Table~\ref{tab:qa_main}): +16.1 percentage points on \texttt{choice\_action}, +19.1 on \texttt{gripper}, +9.5 on \texttt{subgoal}, and +3.5 on \texttt{task\_goal}, for an overall 47.2\% against a pooled 34.6\% chance rate. Figure~\ref{fig:llm_example} illustrates the choice-scoring mechanism on an example \texttt{subgoal} item.

With the language model frozen, Stage~1 clears chance substantially only on \texttt{gripper} (+12.4 points), with the other three question types sitting at their floors: a frozen language model, given only a trained projector, can read instantaneous state out of the weight-tokens. Unfreezing changes this, as fine-tuning multiplies the \texttt{choice\_action} gain roughly $8\times$ and doubles \texttt{subgoal}'s. Progress information is present in the tokens, but consuming it requires reshaping the language model itself, consistent with the findings of LLaNA~\cite{amaduzzi2024llanalargelanguagenerf}.

\texttt{task\_goal} is the most demanding label in the study and carries the narrowest margin (+3.5 points, $\kappa = 0.309$). Under intra-task distractor pooling, every candidate is a near-duplicate goal instance from the same task, differing in a single attribute such as a color or a repetition count, so only instance-level episode content can separate them. Across question types, Cohen's $\kappa$ of 0.31 to 0.44 indicates fair to moderate chance-corrected agreement, with the pooled Stage~2 value of 0.446 approaching the moderate band of~\cite{PMID:843571}.

\begin{table}[t]
\centering
\caption{End-to-end QA performance (meta-learned initialization, pixel-only fitting). Debiased accuracy (mean $\pm$ std across 2 seeds) after subtracting the LLM prior NLL (Eq.~\ref{eq:debias}); chance is the per-item random-guess baseline; gain is Stage~2 accuracy minus chance in percentage points.}
\label{tab:qa_main}
\resizebox{\columnwidth}{!}{%
\begin{tabular}{l c c c c c}
\toprule
\textbf{Question Type} & \textbf{Chance (\%)} & \textbf{Stage 1 (\%)} & \textbf{Stage 2 (\%)} & \textbf{Gain (pp)} & \textbf{$\kappa$} \\
\midrule
\texttt{choice\_action} & 32.0 & $33.9 \pm 0.6$ & $\mathbf{48.1 \pm 2.8}$ & \textbf{+16.1} & 0.444 \\
\texttt{gripper}        & 50.0 & $62.4 \pm 8.2$ & $\mathbf{69.1 \pm 4.5}$ & +19.1 & 0.371 \\
\texttt{subgoal}        & 25.0 & $29.2 \pm 0.9$ & $\mathbf{34.5 \pm 2.7}$ & +9.5 & 0.326 \\
\texttt{task\_goal}     & 29.5 & $29.1 \pm 0.6$ & $\mathbf{33.0 \pm 0.7}$ & +3.5 & 0.309 \\
\midrule
\textbf{Overall}        & 34.6 & $39.6 \pm 1.8$ & $\mathbf{47.2 \pm 2.6}$ & \textbf{+12.6} & \textbf{0.446} \\
\bottomrule
\end{tabular}%
}
\end{table}

\subsection{Memory-Language-Action Decoder}
\label{sec:INR2action}

\begin{figure*}[htbp]
    \centering
    \includegraphics[width=0.95\linewidth]{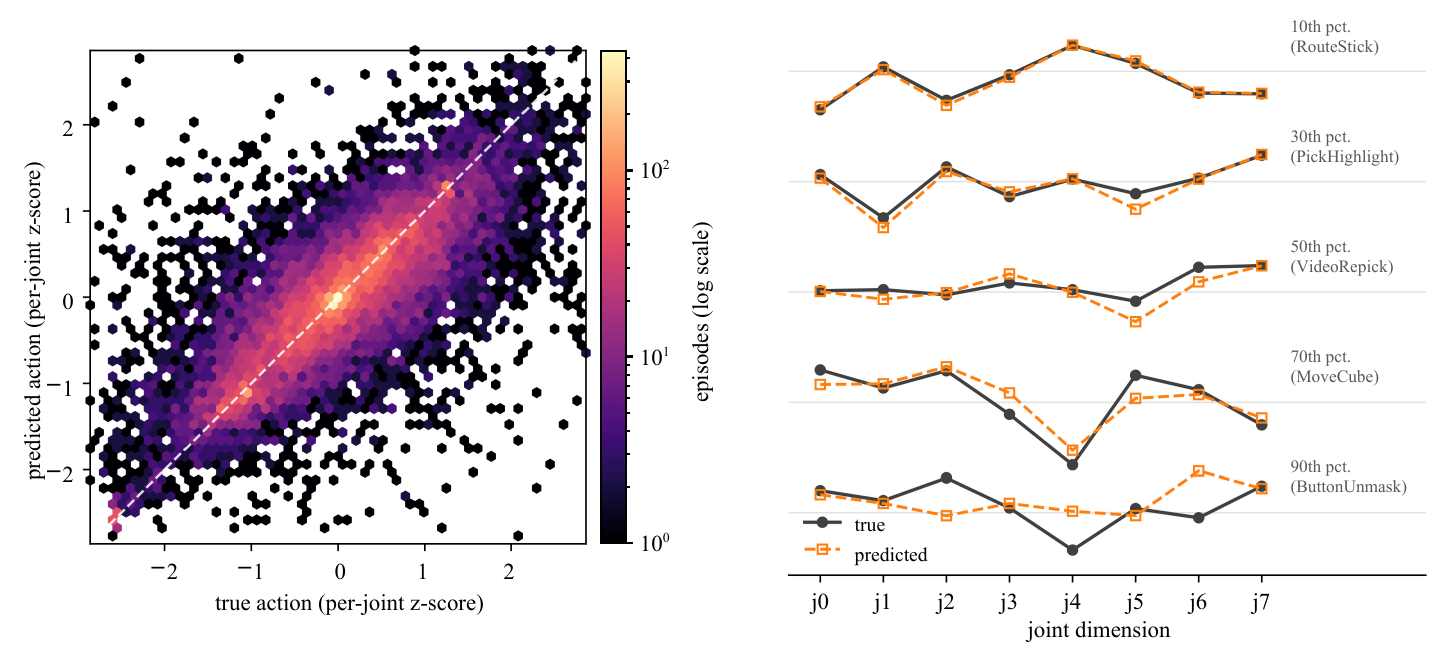}
    \caption{\textbf{Decoding actions from weight-space memory.} Left: predicted against true actions pooled over the seven continuous joints, each z-scored, with density on a log scale; pooled $R^2$ is 0.53. Right: full predicted profiles (orange) against ground truth (black) for episodes at fixed percentiles of mean absolute error, selected by rank rather than by hand.}
    \label{fig:mla_hero}
\end{figure*}

The QA experiments read the weight-space memory retrospectively. This section shows that the same 
$\Delta$ also conditions what the robot does next, and does so against a continuous target, which measures how tightly the memory constrains the action rather than only whether the relevant information is present. The memory is projected into action space through the architecture of Section III-C: a vision-language-action backbone whose vision input is replaced by $\Delta$.

The target is RoboMME's \texttt{joint\_action}, the absolute joint-space command holding the seven target joint angles of the arm and a binary gripper directive, read out at the cut frame as the configuration commanded at the point where the INR's observation ends. This is behavior cloning at a single decision point and, unlike deployed VLA systems evaluated by closed-loop rollout~\cite{brohan2023rt2visionlanguageactionmodelstransfer}, necessarily offline: the input does not exist until the episode is over, so what is measured is action decoding from episodic memory rather than policy performance.

With the backbone frozen, the projector and action head alone reach a mean $R^2$ of $0.48$ over the eight dimensions; Stage~2 fine-tuning raises this to $0.51$ (bootstrap 95\% intervals $[0.459, 0.509]$ and $[0.488, 0.539]$), with the improvement holding in every dimension individually and a mean absolute error of roughly $0.15$ radians. Figure~\ref{fig:mla_hero} presents this performance: pooled over the continuous joints, prediction density concentrates along the identity line, and complete per-episode action profiles track the ground truth closely, with episodes up to the 90th percentile of prediction error recovering the coarse shape of the true action. The gripper command takes only the values $\pm 1$, and thresholding its continuous output yields 81.4\% sign accuracy, exceeding the Stage~2 QA accuracy of 69.1\%.

The controls tie this performance to the episode. Substituting Gaussian noise for $\Delta$ before projection collapses the mean test $R^2$ from $+0.48$ to $-0.39$ for Stage~1 and from $+0.51$ to $-0.55$ for Stage~2, driving every individual dimension below zero: the trained system conditions on the token span, producing wrong actions when the tokens are uninformative rather than retreating to the marginal mean. Recomputed within each task, where the constant instruction can contribute nothing, mean $R^2$ remains positive in all sixteen tasks, spanning roughly $0.19$ (InsertPeg) to $0.80$ (StopCube), so the pooled result is not an artifact of the model separating tasks by their instructions. The unmasking tasks decode best and the fine-manipulation and ordering tasks worst, a plausible reading being that $\Delta$ encodes coarse scene and pose structure more faithfully than fine manipulation detail, a pattern also visible in the QA results. Section~\ref{sec:qa_experiments} showed the weight-space memory can be recalled, and this section shows it can be acted on.

\subsection{Ablation Studies}
\label{sec:ablations}

\subsubsection{Varying INR Initialization and Supervision}
Six INR configurations (``flavors'') are fit per cut, varying two independent factors: the initialization (the meta-learned $\theta$ versus independent random weights) and the supervisory signal used during fitting (pixel reconstruction, self-supervised DINO features~\cite{siméoni2025dinov3}, CLIP features~\cite{radford2021learningtransferablevisualmodels}, or combinations), as specified in Table~\ref{tab:ablation}. Holding one factor fixed while the other varies isolates whether gains come from meta-initialization, auxiliary semantic supervision, or both. For flavors with auxiliary heads, linear DINO and CLIP heads over a pooled feature map are supervised with cosine losses against targets from frozen DINOv3 ViT-B/14 and CLIP ViT-L/14 encoders, extracted once per episode.

\begin{table*}[htbp]
\centering
\resizebox{0.8\textwidth}{!}{%
\begin{tabular}{c l cccc c c}
\toprule
& & \multicolumn{4}{c}{\textbf{Accuracy by Question Type (\%)}} & & \\
\cmidrule(lr){3-6}
\textbf{Flavor} & \textbf{Config.} & \texttt{choice\_action} & \texttt{gripper} & \texttt{subgoal} & \texttt{task\_goal} & \textbf{Overall (\%)} & \textbf{$\Delta_{\text{noise}}$ (pp)} \\
& & \textit{(32.0)} & \textit{(50.0)} & \textit{(25.0)} & \textit{(29.5)} & \textit{(34.6)} & \\
\midrule
1 & R, P & $41.5 \pm 0.7$ & $50.2 \pm 0.5$ & $28.6 \pm 1.3$ & $29.4 \pm 0.4$ & $38.0 \pm 0.5$ & $0.4 \pm 0.1$ \\
\textbf{2} & \textbf{M, P} & $\mathbf{48.1 \pm 2.8}$ & $\mathbf{69.1 \pm 4.5}$ & $34.5 \pm 2.7$ & $33.0 \pm 0.7$ & $\mathbf{47.2 \pm 2.6}$ & $\mathbf{11.4 \pm 0.8}$ \\
3 & M, D & $37.6 \pm 0.6$ & $52.5 \pm 1.8$ & $25.3 \pm 1.0$ & $30.2 \pm 1.9$ & $37.0 \pm 0.8$ & $2.5 \pm 0.8$ \\
4 & M, P+D & $46.1 \pm 4.7$ & $65.6 \pm 0.9$ & $31.6 \pm 2.0$ & $32.8 \pm 0.6$ & $45.0 \pm 2.1$ & $9.7 \pm 1.8$ \\
5 & R, P+D & $40.6 \pm 0.8$ & $51.6 \pm 1.2$ & $27.4 \pm 1.1$ & $31.5 \pm 2.1$ & $38.3 \pm 0.9$ & $1.9 \pm 0.1$ \\
6 & M, P+D+C & $44.7 \pm 3.1$ & $66.2 \pm 1.8$ & $\mathbf{36.2 \pm 2.6}$ & $\mathbf{33.4 \pm 0.1}$ & $46.1 \pm 0.5$ & $8.4 \pm 1.8$ \\
\bottomrule
\end{tabular}%
}
\caption{Ablation over INR configurations, Stage 2 debiased accuracy on the test split, mean $\pm$ std over 2 seeds. Init. is meta-learned (M) or random (R); heads are pixel (P), DINO (D), and CLIP (C). $\Delta$noise is the overall accuracy lost when $\Delta$ is replaced by matched Gaussian noise.}
\label{tab:ablation}
\end{table*}

Initialization is the clearer of the two factors (Table~\ref{tab:ablation}). Flavor 2 exceeds its random-origin counterpart Flavor 1 by +9.2 percentage points overall, with the widest separation on \texttt{gripper}, where random-origin configurations sit at their 50\% floor and meta-initialized ones reach 66 to 69\%. The effects of supervision are more modest and run largely in one direction. Flavor 3, meta-initialized but fitted to DINO features without pixel reconstruction, is the weakest configuration in the table at 37.0\%, below even random-origin Flavor 1: a shared origin without reconstruction is worth less than reconstruction without a shared origin. Flavor 4 trails its DINO-free counterpart Flavor 2 by 2.2 points, and Flavor~5 tracks Flavor 1 within a point on every question type, leaving initialization as the operative factor at this scale. Flavor 6 is the one exception, trailing Flavor 2 overall yet posting the best \texttt{subgoal} and \texttt{task\_goal} values in the table, the two question types whose answers require fine-detail natural-language description such as cube color and count, which is at least suggestive given that CLIP's image embedding is trained against language. For Flavor 2, per-suite accuracy spans a narrow band of roughly 44 to 52\%, so temporal, spatial, object, and procedural memory all contribute to the pooled result rather than one tractable suite carrying it.

\subsubsection{Matched Noise Control}
The noise control of Section~\ref{sec:experiments} separates the meta-initialized and random-origin configurations more sharply than accuracy alone ($\Delta_{\text{noise}}$ column of Table~\ref{tab:ablation}). For Flavor~2, replacing $\Delta$ with noise negatively affects most of the measured performance: overall accuracy falls from 47.2\% to 35.9\%, \texttt{choice\_action} loses 18.5 points and lands below its chance baseline, and every question type ends at or near its per-item chance level. Flavors~4 and~6 behave alike, losing 9.7 and 8.4 points overall under the same substitution. Flavor~1 does not move (38.0\% with real $\Delta$'s, 37.7\% with noise): its apparent +9.5-point margin on \texttt{choice\_action} survives the removal of all episode information, and must therefore derive from question and candidate statistics learned during fine-tuning rather than from anything the random-origin weights encode. Flavors~3 and~5 likewise move by only 2.5 and 1.9 points. This verifies that shared origin is what makes the weight-tokens informative about episode content at all.

\section{Conclusion}

Experimental results establish projected INR weight-delta tokens as a queryable episode representation: a language model answers memory-dependent questions from the weight-tokens alone, an action decoder predicts the robot's next joint command from the same tokens, and meta-learned parameter origins are shown to be necessary for such information to be decodable. The weight delta $\Delta$ thereby functions as a persistent and individually addressable memory object, a property no activation-bound representation offers, and one that complements continually overwritten parametric memories such as Titans~\cite{behrouz2024titanslearningmemorizetest} wherever individual episodes must remain separable. For robotics, where RoboMME identifies memory as a binding constraint on generalist policies~\cite{dai2026robommebenchmarkingunderstandingmemory}, our results suggest memory representations, rather than frame retention alone, as a design axis for vision-language-action systems~\cite{wen2025tinyvlafastdataefficientvisionlanguageaction}.

There are several limitations. All accuracies rest on two training seeds and a limited corpus of source episodes, so differences of one to three points between adjacent configurations should not be over-read. Scoring a closed candidate set by likelihood establishes discrimination, not free-form generation, and the memory construction is offline, with no external baseline such as a vision-language model given the frames directly. Scaling INR capacity and resolution, evaluating open-ended generation, and moving the INR fit and language model updates online as test-time training in the sense of Titans \cite{behrouz2024titanslearningmemorizetest} are delegated to future work.

\section*{ACKNOWLEDGMENT}
\noindent
SWP is a founder of Matta Labs, and CM is an employee; both hold equity in the company, which develops AI systems for manufacturing. The remaining author declares no competing interests.

\bibliographystyle{IEEEtran}
\bibliography{references}

\end{document}